\pdfoutput=1

\documentclass[11pt]{article}

\usepackage[]{EMNLP2022}

\usepackage{times}
\usepackage{latexsym}

\usepackage[T1]{fontenc}

\usepackage[utf8]{inputenc}

\usepackage{microtype}

\usepackage{graphicx} 
\usepackage{ulem}
\usepackage{color}

\usepackage{amsmath}

\title{Lexicon-injected Semantic Parsing for Task-Oriented Dialog}

\author{Xiaojun~Meng$^{1}$\thanks{$^*$~Email: xiaojun.meng@huawei.com}, ~Wenlin~Dai$^{1,2}$\thanks{$\dagger$ Work is done at the internship of Noah's Ark Lab, Huawei Technologies.}, ~Yasheng~Wang$^{1}$, Baojun~Wang$^{1}$, \\ \textbf{Zhiyong~Wu}$^{2}$, \textbf{Xin~Jiang}$^{1}$, \textbf{Qun~Liu}$^{1}$\\
  \\
 $^1$ Noah’s Ark Lab, Huawei Technologies \\
 $^2$ Shenzhen International Graduate School, Tsinghua University, Shenzhen, China
 }

\begin{document}
\normalem
\maketitle
\begin{abstract}
Recently, semantic parsing using hierarchical representations for dialog systems has captured substantial attention. Task-Oriented Parse (TOP), a tree representation with intents and slots as labels of nested tree nodes, has been proposed for parsing user utterances. Previous TOP parsing methods are limited on tackling unseen dynamic slot values (\emph{e.g.,} new songs and locations added), which is an urgent matter for real dialog systems. To mitigate this issue, we 
first propose a novel span-splitting representation for span-based parser that outperforms existing methods. Then we present a novel lexicon-injected semantic parser, which collects slot labels of tree representation as a lexicon, and injects lexical features to the span representation of parser. An additional slot disambiguation technique is involved to remove inappropriate span match occurrences from the lexicon. Our best parser produces a new state-of-the-art result (\textbf{87.62\%}) on the TOP dataset, and demonstrates its adaptability to frequently updated slot lexicon entries in real task-oriented dialog, with no need of retraining.
\end{abstract}

\section{Introduction}
Traditional intent classification and slot-filling methods \citep{tur2011spoken, louvan-magnini-2020} are widely used in dialog systems to parse task-oriented utterances, however, whose representation is flat and limited. It is usually composed of a single intent per utterance and at most one slot label per token. It's difficult for such a flat representation to handle compositional and nested queries. For instance, an intent is included inside a slot, which is common in commercial conversational systems with multiple backend services \citep{pasupat2019span}. 

\begin{figure}
    \centering
    \includegraphics[scale=0.4]{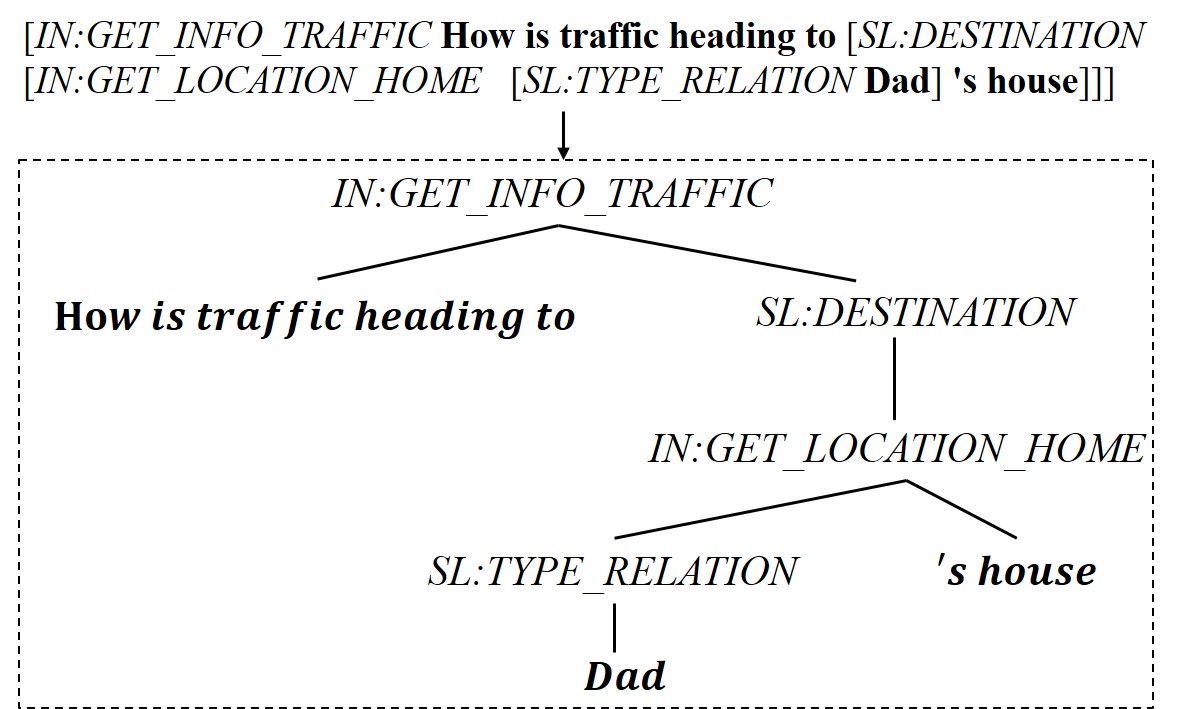}
    \caption{Hierarchical representation of an utterance from TOP dataset (IN: = intent; SL: = slot).}
    \label{fig-1}
\end{figure}

To overcome the limitation of classical intent-slot frameworks, \citet{gupta2018semantic} proposed a hierarchical Task-Oriented Parsing (TOP) representation allowing slots to contain nested intents, as presented in Figure \ref{fig-1}. They empirically show the TOP representation is expressive enough to model the vast majority of human-generated complex queries in given domains. Furthermore, this representation is easier to annotate and parse than alternatives such as logical forms. 

A large amount of semantic parsers \citep{pasupat2019span, einolghozati2019improving, sota, rongali2020don} are designed on this task-oriented representation. In particular, given an utterance $x$ = ($x_0,  x_1, \dots, x_{n-1}$) with $n$ tokens, existing parsers are able to parse $x$ into the tree representation, as illustrated in Figure \ref{fig-1}. However, dialogues in TOP dataset are constrained by a \emph{fixed domain ontology}, which essentially describes all the possible values that each predefined slot can take \citep{xu-hu-2018}. In real dialogs, some slot values may not appear in the train set, and are called unseen slot values \citep{xu-hu-2018}.

Therefore, how to effectively handle unseen slot values in task-oriented parsers is very important. In real dialog system, a couple of slot lexicon tables are often used to guide slot-filling. 
For instance, restaurants and locations nearby are known in advance and updated frequently in the lexicon table to guide phone assistants. However, due to the issue of \emph{fixed domain ontology}, it usually requires retraining for existing parsers to tackle unseen and dynamic lexical entries, which is time-consuming and computationally expensive.

Lexicon-based methods have demonstrated capabilities in various tasks such as sentiment analysis \citep{taboada2011lexicon} and NER \citep{li2020flat}. The lexicon-based method is up-and-coming for task-oriented semantic parsing with the concept of \emph{slot}, since slot values are easily and meaningfully collected as a \emph{slot lexicon}. We are thus motivated to design a lexicon-injected parser, that is capable of leveraging lexicon resources to improve parsing utterances and its easy adaptation to tackle unseen and dynamic lexical entries. In this adaptation, retraining is not required, making it less time-consuming for use in real dialog systems.

Inspired by \citet{pasupat2019span}, we propose a span-based and lexicon-injected parser that embeds each span token as a vector to predict the labels of the tree nodes covering this span. This label predication uses independent scoring of span representation, which is further enriched via either the splitting feature from its child nodes (section \ref{split-feature}), or the valuable lexical feature from matching utterances to a collected slot lexicon (section \ref{lexicon-injected}). In particular, we tag each matched span as the slot category it belongs to, and propose a slot disambiguation model to remove slot mismatch occurrences. Our contributions in this paper are three-folds:
\begin{enumerate}
\item We present a span-based semantic parser that sets state of the art in parsing TOP representation, and incorporates splitting feature to represent each parent span as a root of subtree.
\item We propose a novel lexicon-injected method to further improve compositional semantic parsing, via the use of slot disambiguation technique to remove slot mismatch occurrences in the lexicon.
\item We explore the potentials of adapting lexicon-injected parsers to parse frequently updated lexical entries in real-time task-oriented dialog, via the human-driven adaptation of altering the lexicon, with no need of time-consuming retraining.
\end{enumerate}

\section{Related Work}
\paragraph{Language understanding for task-oriented dialog.}  Most neural methods \citep{louvan-magnini-2020} on slot filling and intent classification for task-oriented dialogue systems are using the traditional flat representation. The classical intent-slot framework is to identify a single user intent and then fill the relevant slots. For instance, a large amount of work tackles on the ATIS \citep{price1990evaluation, zhu2017encoder} and DSTC \citep{williams2016dialog} datasets. Among them, \citet{xu-hu-2018} highlight a practical yet rarely discussed problem in DSTC, namely handling unknown slot values.

\paragraph{Semantic parsing.} Traditional methods of semantic parsing work on creating the logical form or dependency graph for utterances \citep{berant2013semantic, van2018exploring}, which is expressive but brings a significant barrier to learning and annotation. \citet{gupta2018semantic} propose
a tree hierarchical and easily annotated representation where intents and slots are nested that is able to tackle compositional utterances. Recent approaches on this new hierarchical representation are span-based parser \citep{pasupat2019span, herzig2020span}; ensemble/re-tanking model using RNNGs \citep{einolghozati2019improving}; sequence-to-sequence neural networks augmented with pointer-generator architecture \citep{jia-liang-2016, sota, rongali2020don}.

\paragraph{Lexicon-based methods.} Lexicon-based methods have demonstrated the superiority in various NLP tasks. \citet{taboada2011lexicon} present the lexicon-based SO-CAL to extract sentiment from text, by using a dictionary (e.g., \emph{opinion lexicon}) of opinion words to identify and determine sentiment orientation. They show that SO-CAL's performance is consistent across domains and on completely unseen data. \citet{li2020flat} introduce FLAT: a flat-lattice transformer to incorporate lexicon information for Chinese NER. It uses a lexicon to match the potential entity entries in sentences. We believe this lexicon enhanced method can be further applied to semantic parsing. There are several lexical resources suitable for semantic parsing, built with extensive human efforts in years, including VerbNet \citep{kipper2000class}, WordNet \citep{miller1995wordnet}, and combined resources \citep{shi2005putting}. We are unaware of any semantic parser using a slot lexicon for parsing task-oriented dialog.

\section{Model}

\begin{figure}
    \centering
    \includegraphics[scale=0.5]{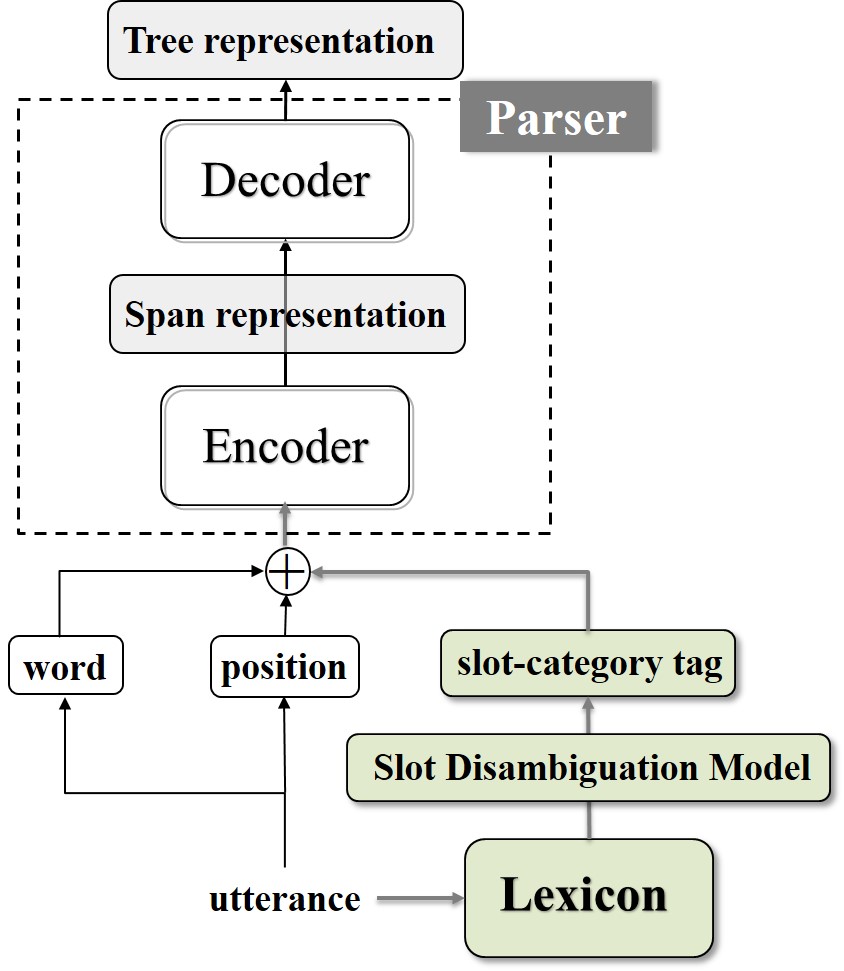}
    \caption{Encoder-Decoder architecture of our span-based parser. The light green part is our lexicon-injected method, which tags each span matched in the lexicon as the slot category it belongs to. The slot disambiguation is used to remove inappropriate slot-match occurrences in the lexicon, as further shown in Figure \ref{SD-demo}.}
    \label{fig-2}
\end{figure}

Our base model is a span-based parser as firstly used in \citet{stern2017minimal} for constituency parsing. A constituency tree over an utterance is a collection of labeled spans. Given an utterance, the task is to predict a parse tree. Span $(i, j)$ corresponds to the constituent that is located between position $i$ and $j$ in an utterance. The label of a span is either an intent or a slot (prefixed with \texttt{IN:} or \texttt{SL:} in Figure \ref{fig-1}).
Our parser follows an encoder-decoder architecture (Figure \ref{fig-2}). The encoder is a two-layer transformer \citep{vaswani2017attention} and the decoder is a chart parser borrowed from \citet{stern2017minimal}.

\subsection{Encoder}
\label{encoder}
We first represent each word $x_{i}$ using two pieces of information: an external context-aware word representation $w_{i}$ and a learned position embedding, where every position $i \in 0,1, \dots, n-1$ is associated with a vector $p_{i}$. We concatenate these two embeddings to generate a representation of a word:
$$x_i = [w_i; p_i]$$
\paragraph{Span representation.} To get the representation of span $(i, j)$ in the utterance, we first introduce a notation of \emph{boundary} between each consecutive words $(x_{i-1},x_{i})$ in the utterance. We feed the entire utterance $x$ = ($x_1, \dots, x_{n}$) into a two-layer Transformer to calculate the \emph{boundary representation} as $f_{i-1, i}$. Specifically, given the output $h_i$ from the encoder for word $x_i$, we split it into two halves $[h^1_i;h^2_i]$ and define $f_{i-1, i} = [h^2_{i-1};h^1_i]$, which means each boundary representation is decided by the left and right context together. In this way, the dimension of $f_{i-1, i}$ is the same as $h_i$ (see Appendix \ref{appendix_A}).

For an utterance $x$ with length as $n$, we are able to calculate the context-aware boundary representations $(f_{0, 1}, f_{1, 2}, \dots, f_{n-2, n-1})$ with size of $n-1$. Notice that the beginning and ending tokens are special tokens such as \texttt{[CLS]} and \texttt{[SEP]} used in \citet{bert}.

We define the representation of span $(i, j)$ as:
$$ r_{i, j} = f_{j, j+1} - f_{i-1, i}$$
which uses the left and right boundaries to decide the in-between span representation.

\subsection{Decoder}

We use the chart parser of \citet{stern2017minimal} with additional modifications in \citet{gaddy2018s} and \citet{kitaev2018constituency}. Our parser assigns a score function $s(T)$ to each mapping $T$, which can be regarded as a tree shown in Figure~\ref{fig-1}. It decomposes as:
$$ s(T) = \sum_{T: (i, j) \rightarrow l}s(i, j, l)$$
where $s(i, j, l)$ is a real-valued score for the span $(i, j)$ that has label $l$. This span-label scoring function is implemented as a one-layer feedforward network, taking as input the span representation $r_{i, j}$ and producing as output a vector of label scores:
\begin{align}\label{s_label_1}
s_{label}(i,j) &= V\mathtt{ReLU}(W r_{i, j} + b),\\
s(i,j,l) &= [s_{label}(i,j)]_{l}
\end{align}

Let $k$ denote the index of the splitting point where span $(i, j)$ is divided into left and right child nodes, we define $s^{*}(i,j)$ as the score of the best subtree spanning $(i, j)$:
\begin{equation}
\label{s_best}
\begin{split}
s^*(i, j) = &\mathop{\max}_{l, k}[s(i, j, l) \\
&+ s^*(i, k-1) + s^*(k, j)]
\end{split}
\end{equation}

For the simplest case, the leaf node is a single word of an utterance. We calculate $s^{*}(i,i+1)$ as:
\begin{equation}
\label{s_i_i+1}
\begin{split}
s^*(i, i+1) = &\mathop{\max}_{l}[s(i, i+1, l)]
\end{split}
\end{equation}

To parse the full utterance $x$, we compute $s^*(1, n)$ and then traverse backward to recover the tree achieving that score. Therefore, the optimal inference tree $T^*$ can be found efficiently using a CKY-style bottom-up inference algorithm:
$$T^* = \mathop{\arg}\max_{T}\ s(T)$$

We follow the common approaches \citep{stern2017minimal, pasupat2019span} of handling the unary chain by a collapsed entry, and the $N$-ary trees by binarizing them and introducing a dummy label. In terms of handling unary chain, the set of labels includes a collapsed entry for each unary chain in the training set. For an instance in figure \ref{fig-1}, the set (\emph{SL:DESTINATION}, \emph{IN:GET\_LOCATION\_HOME}) is an unary chain for the span ``\texttt{Dad 's house}''. A TOP tree is often $N$-ary since it has spare constituents. The model hands $N$-ary trees by binarizing them and introducing a dummy label $\emptyset$ to nodes created during binarization, with the property that: $s(i,j,\emptyset) = 0$.


\subsection{Training Objective}
To train the model, we use the margin loss as described in previous works \citep{stern2017minimal}. Given the correct tree $\hat{T}$ and predicted tree $T$, the model is trained to satisfy the margin constraints:
\begin{equation}
s(\hat{T}) \ge s(T) + \theta(\hat{T}, T)
\end{equation}

Here $\theta$ is a distance function that measures the similarity of labeled spans between the prediction and the correct Tree. The margin loss is calculated as:
\begin{equation}
\max[0, -s(\hat{T}) + \max[s(T) + \theta(\hat{T}, T)]]
\end{equation}

\subsection{Span-splitting Representation}
\label{split-feature}
The majority of parsing models score each span independently, which is sub-optimal. To mitigate the independence assumption between node labels, \citet{pasupat2019span} introduce \emph{edge scores}: concatenates child node’s span embedding with its additional label embedding as input, and thus model the conditional distribution over all possible labels of its parent node. Via this method, \emph{edge scores} connect a parent node label to its child node label.

However, a parent node is still unaware of the split point, where this parent span is divided into two child spans. Notice that in equation \ref{s_best}, the label scoring of parent span $s(i, j, l)$ is unaware of the underlying split point $k$. We propose a simple but effective way of incorporating this splitting feature into the decision, without introducing a new embedding like edge scores. We define a new \textbf{\emph{span-splitting representation}} $\hat{r}_{i, j}$, that adds the boundary representation at the splitting point to the parent:
\begin{equation}
\hat{r}_{i, j} = r_{i, j} + f_{k^*-1, k^*}
\end{equation}

where $k^*$ is the best splitting point dynamically computed as follows, dividing the parent span into two separate spans $(i, k-1)$ and $(k, j)$:
$$k^* = \mathop{\arg}\max_{k} [s^*(i, k-1) + s^*(k, j)]$$

With this new representation $r$, each span becomes dynamic and more expressive as a root of the subtree it contains. Even for the same span, this new span representation with added splitting feature, can be different if its subtree structure varies. Notice that this modification has no effect on the decoder since we are using the dynamic programming algorithm to perform bottom-up CKY decoding.

\section{Lexicon-injected Method}
\label{lexicon-injected}
Motivated by existing works \citep{taboada2011lexicon, li2020flat} on leveraging lexicon resources to improve related task performances, we believe collecting a slot lexicon table beforehand and injecting the related lexical features to parsing models would also help. Meanwhile, using lexicon information equips the model with the ability of easy adaptation to detect and recognize unseen slot values, which previous TOP parsers are unable to deal with.

\subsection{Lexicon-injected model}
\label{lexion_model}
\begin{figure}
    \centering
    \includegraphics[scale=0.4]{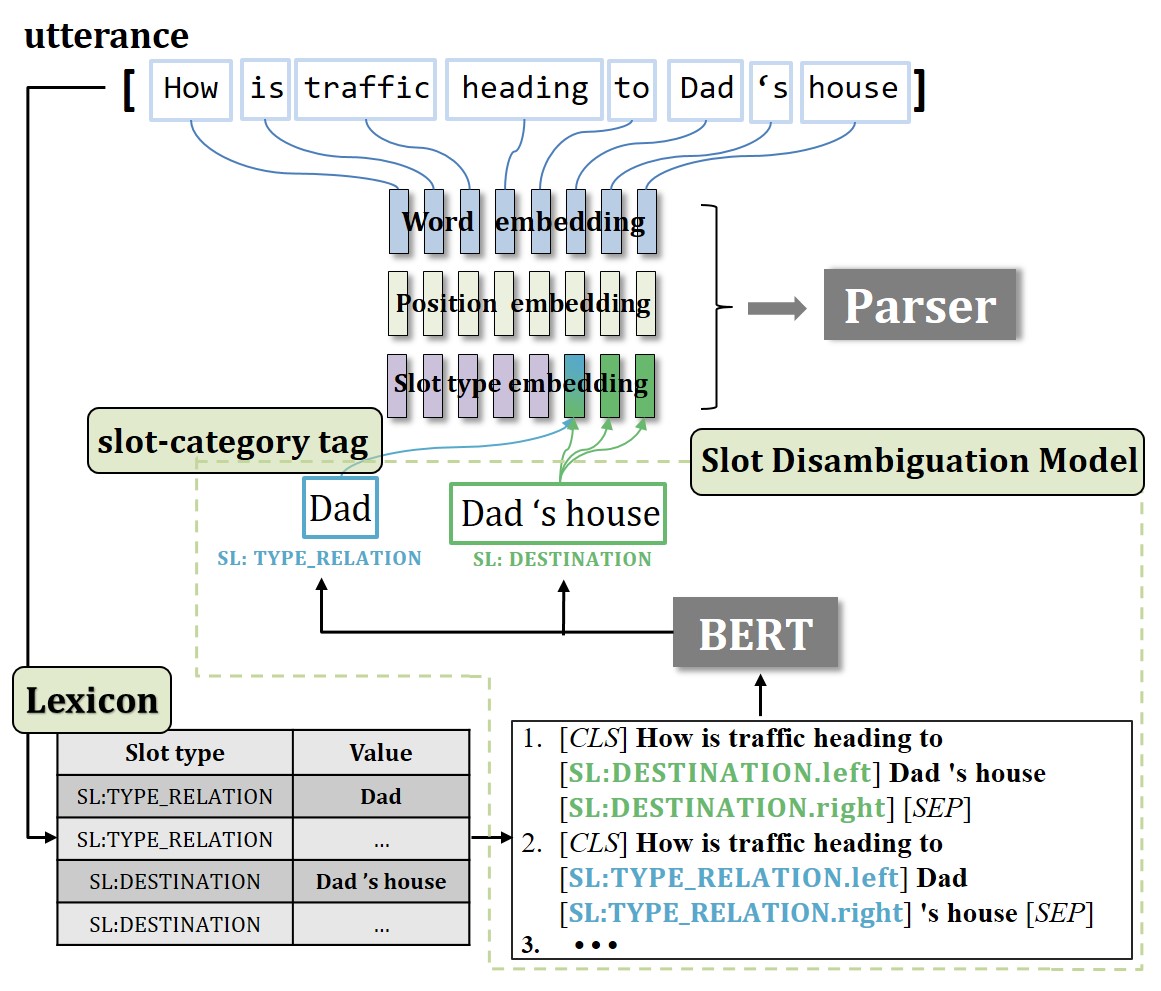}
    \caption{Demonstration of our slot disambiguation technique to inject lexical features to the input embedding for lexicon-injected parsers.}
    \label{SD-demo}
\end{figure}

We first collect a lexicon table: in the TOP train set, there are 36 distinct slot categories, and 20059 slot values (15238 are unique), which means each slot category has 557 ones on average. The lexicon is only built from train set since we have no clue of test set. 
When a span of utterance is included in some slot-category of the lexicon, we tag each word in the span as this slot-category. Therefore, as in Figure~\ref{SD-demo}, one word may have multiple tags that are added together. For those words without any tag, we use an out-of-category tag $t_{o}$ instead. 

The 36 slot categories plus $t_{o}$ compose a vocabulary with 37 learnable embeddings. We concatenate this randomly initialized slot-category embedding $q_i$ to the previous word embedding:
$$x_{i} = [w_{i}; p_{i}; q_{i}]$$
Since one word may match multiple slot entries in the lexicon table, we add all matched slot-category embeddings together as the final $q_{i}$. For word not appearing in any slot category, $q_{i}$ is imputed as a special vector $q_{o}$. The decoder remains the same.

\paragraph{Generalized representation.}
In addition, to empower the slot-category embedding to affect parsing tasks, we propose to replace $w_{i}$ with $q_{i}$ when this word appears in the lexicon as shown in equation (\ref{generalized_equation}). In this generalized representation, we abandon the original word embeddings and enforce the slot-category embedding to take control of representing spans from the lexicon. Via this replacement, we also aim to make the model more informative of slot categories and less constrained to specific slot values, and thus it is able to model newly added unseen slot values the same as seen ones in the lexicon.
\begin{align}
\label{generalized_equation}
x_{i} = \begin{cases}
[w_{i}; p_{i}; q_{i}], & t_{i} = t_{o} \\
[q_{i}; p_{i}; q_{i}], & \mathrm{otherwise.} \\
\end{cases}
\end{align}

However, slot categories are often overlapping on words and spans.
As in Table \ref{sd-example}, given the context of utterances, a word or span may mismatch slot entries in the lexicon.
This mismatch brings unexpected noises to the $q_{i}$, especially in the case of nested hierarchical representation like the TOP dataset. Therefore, we further propose a disambiguation technique that aims to remove inappropriate slot matched occurrences.

\subsection{Slot Disambiguation}

We regard this lexicon-mismatch problem as a sequence binary classification with the input of utterance, slot category and slot position, and output whether this slot match occurrence is correct or not in the given context.

For each utterance, the slot category associated with the right position as annotated in the parse tree is labeled as \emph{True} (positive sample); otherwise, it’s a mismatch labeled as \emph{False} (negative sample). For instance of Table \ref{sd-example}, this utterance has multiple matched entries in the lexicon, and only positive samples are consistent with its ground-truth parse tree. We use all these samples for training the disambiguation model.
Given the large number of slot entries in the lexicon, the mismatched negative samples are often more than positive samples.
Therefore, we also try to use different sampling strategies to balance the number of negative and positive ones. We haven't noticed any significant resulting difference in performance.

\begin{table}
\centering
\small
\texttt{How$_1$ is$_2$ traffic$_3$ heading$_4$ to$_5$ Dad$_6$ 's$_7$ house$_8$}\\
\vspace{3mm}
\begin{tabular}{l l l}
\emph{SL:DESTINATION} & \emph{6:8} & \emph{True} \\
\emph{SL:TYPE\_RELATION} & \emph{6:6} & \emph{True} \\
\emph{SL:CONTACT} & \emph{6:6} & \emph{False} \\
\emph{SL:DESTINATION} & \emph{8:8} & \emph{False} \\
\emph{SL:SEARCH\_RADIUS} & \emph{5:5} & \emph{False} \\
\dots \\
\end{tabular}
\caption{\label{sd-example} Example of labeled data for training slot disambiguation model. Columns are slot category, slot position in the utterance and binary labels.}
\end{table}

In the disambiguation model, we insert the slot category as a token to the left and right boundary of corresponding slot values in the origin utterance (see colorful tokens in Figure~\ref{SD-demo}) and feed them as a whole into the pretrained language model (\emph{e.g.,} BERT-base from \citet{bert}) to perform sequence classification. In details, each slot category composes two unused slot-category tokens: the left and right to the slot value, which means there are 72 new introduced tokens such as ``\emph{[SL:DESTINATION.left]}'' and ``\emph{[SL:DESTINATION.right]}'' in Figure ~\ref{SD-demo}.

The pretrained context-aware word embedding and slot-category token embedding keep updated in the training session. Finally, we use the [\emph{CLS}] hidden state $h_{cls}$ to perform sequence classification on label $c$ as commonly used in \citet{bert}:

$$p(c | h_{cls}) = \mathrm{softmax}(W h_{cls} + b)$$

In the inference, we compare each utterance to the collected lexicon and find all the match occurrences. Then we use this disambiguation model to classify each match occurrence and remove inappropriate ones given the context. This filter aims to largely reduce the noisy tagging when constructing the lexicon-injected input embedding $x_{i}$ for the subsequent parsing model.

\paragraph{Adaptation to unseen values.} 
By updating lexicon with the latest slot values, our lexicon-injected method can be easily adapted to parse unseen slot values from a predefined slot category, with no need of retraining both the slot disambiguation model and the parser. We demonstrate this easy and quick adaptation merit in the Experiments.

\section{Experiments}
\label{experiment}
We evaluate our proposed models on the TOP dataset \citep{gupta2018semantic}, which has a hierarchical representation of intent-slot annotations for utterances in navigation and event domains.~\footnote{\url{http://fb.me/semanticparsingdialog}} All the code of this paper will be made public.
\subsection{Evaluation on TOP Representation}
\paragraph{Baselines.}
Most baselines for this task are well described in \citet{gupta2018semantic, sota}. We only include the most competitive baselines in the Table~\ref{result-table}: the first span-based parser on TOP representation \citep{pasupat2019span}, with an additional improvement of using edge scores to model relations between parent and child labels; the generative model Seq2SeqPtr \citep{rongali2020don} based on the Pointer-Generator architecture to understand user queries; a family of Seq2Seq models (decoupled RoBERTa/BART) \citep{sota} that set state of the art in parsing decoupled TOP representation.

\begin{table}\normalsize
\centering
\begin{tabular}{lrl}
\hline \textbf{Method} & \textbf{Acc} & \textbf{F1} \\ \hline
\textbf{Non-lexicon-injected parser:} &  &  \\
Pasupat & 80.80 & 93.35 \\
Pasupat-edge & 81.80 & 93.63 \\
Decoupled RoBERTa & 84.52 & - \\
Decoupled BART & 87.10 & - \\
Seq2SeqPtr (+BERT) & 83.13 & - \\
Seq2SeqPtr (+RoBERTa) & 86.67 & - \\
Ours (base)\dag & 83.06  & 94.23 \\
Ours (+Split)\dag & 83.97 & 94.55 \\
Ours (+RoBERTa) & 85.77 & 95.24 \\
 &  &  \\
\textbf{Our lexicon-injected parser:} &  &  \\
w/o Slot Disambiguation\dag & 81.83 & 93.87 \\
w/ Slot Disambiguation\dag & 85.63 & 96.13 \\
w/ SD + GR\dag & 86.80 & 96.34 \\
w/ SD + GR + RoBERTa & \textbf{87.62} & \textbf{96.60} \\
\hline
\end{tabular}
\caption{\label{result-table} Comparison of complete match accuracy and labeled bracket F1 of different methods on TOP test set. SD, GR and \dag\ denote slot disambiguation, generalized representation and use BERT-base model.}
\end{table}

\paragraph{Methods.}
We evaluate on a few variants of models: 1) base model with BERT-base \citep{bert} or RoBERTa-base \citep{liu2019roberta} as contextualized word embedding ; 2) adding splitting feature to span representation as $\hat{r}_{i, j}$; 3) the trivial lexicon-injected parser without using slot disambiguation; it means there exists an amount of noisy slot mismatch for $q_{i}$ embedding; 4) the lexicon-inject parser with oracle hints from the ground-truth parse tree; it means there is no slot mismatch (Appendix \ref{appendix_oracle_parser}); 5) the lexicon-injected parser with slot disambiguation technique; For 4) and 5), we use an additional generalized representation to empower the slot-category embedding. Once we use an external pretrained model with a different dimension of word embedding to $w_{i}$, we simply apply a learned one-layer feedforward network to align it. Training details can be found in Appendix \ref{appendix_A}.

\paragraph{Results.}
We report the exact match accuracy and the labeled bracket F1 score as widely measured for the parse tree constituents \citep{black1991procedure}. Our base model with the transformer encoder outperforms span-based baselines in two measurements (\textbf{83.06/94.23\%}), as shown in Table~\ref{result-table}. This might not be a surprising result given baselines \citep{pasupat2019span} are mostly using biLSTMs as sequence embedder. We believe pretrained transformer embedders such as BERT are more semantic expressive than RNN models.

We find that adding split-up information (+Split) to the embedding (\textbf{\emph{span-splitting representation}}) is able to improve the complete match by almost one percent (\textbf{+0.91\%}), which is quite significant for intent classification and slot-filling tasks. The result suggests bottom-up splitting decisions from child spans contribute to improve the parser.

The trivial lexicon-injected parser without using slot disambiguation technique (w/o SD) is not even comparable (\textbf{81.83\%}) to the base model, because utterances highly overlap with the lexicon, which brings an amount of unexpected mismatch to the input embedding. We also evaluate on the oracle parsers that have no mismatch spans via the help of ground-truth dev trees. Details can be found in Appendix \ref{appendix_oracle_parser}. This upper bound result motivates us to come up with a slot disambiguation technique.

Results in Table~\ref{result-table} show that the slot disambiguation technique (w/ SD) is very promising to remove inappropriate span match occurrences, and thus largely improves the downstream parsing accuracy from \textbf{83.06\%} to \textbf{85.63\%}. The classification result of slot disambiguation is presented in the section \ref{slot_disambiguation}. Moreover, the additional generalized representation (SD + GR) brings \textbf{1.17\%} improvement to our parser, which shows the generalized representation makes our parser be more informative of slot categories when tackling related slot values.

\begin{figure}
    \centering
    \includegraphics[scale=0.4]{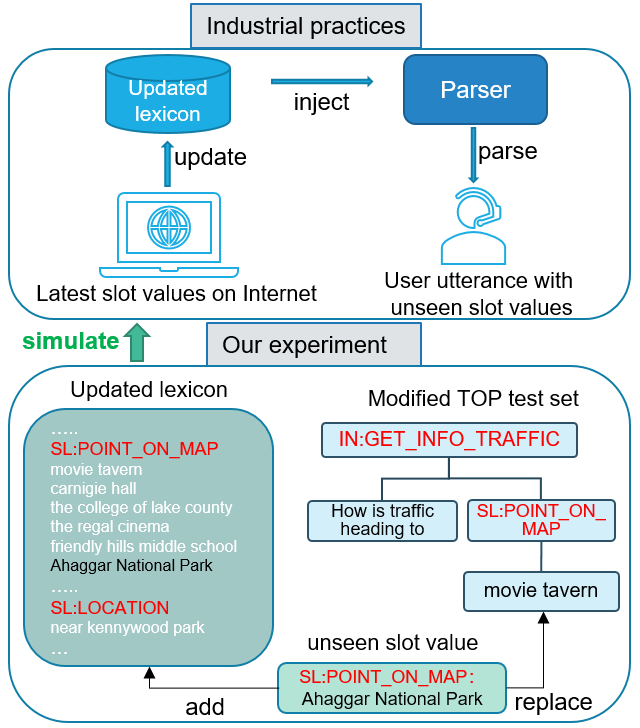}
    \caption{Simulation of the real-world lexicon updating for industrial need. Up-to-date slot values can be imported into the lexicon automatically or manually, such as \emph{POINT\_ON\_MAP} from the Internet map resource.}
    \label{exp_simulation}
\end{figure}

In addition, our base model (+RoBERTa) outperforms decoupled RoBERTa by \textbf{1.25\%} percent, and achieves comparable performances to the modern generative Seq2SeqPtr. 
To our knowledge,the decoupled method is generally not as good as Seq2SeqPtr given the fact that Seq2SeqPtr with RoBERTa achieves a better performance (\textbf{+2\%}) than decoupled RoBERTa. The performance of the decoupled BART \citep{sota} is better while we believe the credit goes to its high-capacity pretrained encoder. We aim to compare those methods in the same encoder setting. 
Overall, our best parser (SD + GR + RoBERTa) achieves the new state of the art (\textbf{87.62\%}), which is even better than decoupled BART and an ensemble of RNNGS \citep{einolghozati2019improving}.

\begin{table} \normalsize
\centering
\begin{tabular}{lrl}
\hline \textbf{Method} & \textbf{Acc} & \textbf{F1} \\ \hline

\textbf{Lexicon-injected parser:} &  &  \\
w/ Slot Disambiguation\dag & 84.80 & 95.92 \\
w/ SD + GR\dag & 86.30 & 96.25 \\
w/ SD + GR + RoBERTa & \textbf{87.27} & \textbf{96.54} \\
\textbf{Non-lexicon-injected parser:} &  &  \\
Pasupat & 70.90 & - \\
Ours (+Regex substitution)\dag & 69.94 & - \\
Ours (+RoBERTa) & 74.35 & 92.10 \\
Seq2SeqPtr (+RoBERTa) & 74.91 & 92.42 \\

\hline
\end{tabular}
\caption{\label{modified-test-result} Comparison of complete match accuracy and labeled bracket F1 of different methods on the modified TOP test set with introducing unseen slot values. SD, GR and \dag\ denote slot disambiguation, generalized representation and using BERT-base respectively.}
\end{table}

\subsection{Adaptation to unseen slot values}
\label{generalization}
Our lexicon-injected method can be used in real parsing dialog for industrial needs as follows. The latest slot values, from predefined slot categories, are always generated. Some of them can be automatically updated to the lexicon via the Internet resource, such as new places and new songs. In addition, some slot values related to a category were originally missing during the training, and are later added to the lexicon. For instance, new names are inserted to the contact lexicon of phone assistants. In no matter which manner, updating the lexicon with a wide coverage of slot values can better guide the model to understand user utterances, which is particularly useful in the industrial system.

To investigate how the lexicon-injected method is able to handle unseen slot values that are missing in the training.
As in Figure~\ref{exp_simulation}, our experiment simulates the real-world scenarios. We update the lexicon table by involving up-to-date slot values that are likely to appear in user utterances. 
In particular, we design five new slot values to two randomly predefined slot categories in the lexicon. 
Meanwhile, we have to evaluate models on utterances that indeed include these new values. 
Therefore, we build a modified test set based on TOP test set,
by randomly selecting these new values and replacing original slots in utterances that belong to the same slot category. This replacement on each utterance randomly happens and we are unaware of which old slot value might or might not be replaced.

We believe this way of updating the lexicon and building a modified test set is simulating the real-world scenario as much as possible. In total, there are 1621 modifications (19.6\%) on 8241 utterances of the test set. Note that we can introduce more modifications when increasing the possibility of replacement operation on each utterance. Examples can be found in Appendix \ref{appendix_test_set}.

\paragraph{Results.} As shown in Table~\ref{modified-test-result}, without re-training any part of the model, our lexicon-injected parsers are still able to achieve comparable performances with before. In particular, the complete match accuracy for the parer (w/ SD + GR) is only reduced by \textbf{0.4\%} (from 87.62\% to 87.27\%). We also evaluate state-of-the-art models on this modified test set. Since the modified test set encompasses a wide range of up-to-date slot values, unfortunately, the performance of existing parsers without re-training drops dramatically (around \textbf{-11\%} for all models).

\begin{figure}
    \centering
    \includegraphics[scale=0.5]{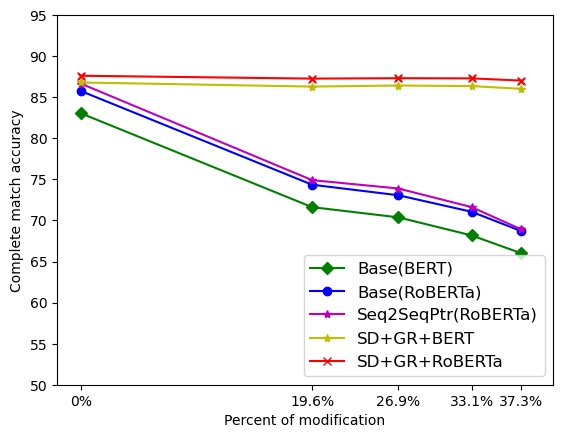}
    \caption{Comparison of complete match accuracy of different methods on varying modified TOP test set.}
    \label{fig-large-scale}
\end{figure}

\paragraph{Regex substitution.} We also consider other potential human hard-coding adaptation such as performing regex substitution on user utterances. In particular, when a user utterance comes, we 
substitute all possible new slot values with random old alternatives from the same slot category. Then we feed utterances to our trained parsers to get the parsed trees, and finally undo substitution on trees to get the final intent and slot results.

However, the performance of human hard-coding adaptation is relatively low due to the overlapped entries of new slot values on user queries. For instance, a new slot value \texttt{"bridge"} is added to \emph{SL:LOCATION}. Unfortunately, \texttt{"bridge"} is also an existing entry in \emph{SL:PATH} and \texttt{"san franciscos bridge"} is in \emph{SL:DESTINATON}. The hard-coding regex rules have no knowledge of context in a query, and thus can bring randomness and noises to parsers.

\paragraph{Retraining.} Retraining existing state-of-the-art models with adding utterances including unseen slot values to the train set, can recover the performance to a similar result in Table~\ref{result-table}. However, such retraining process is time-consuming and unrealistic to perform in real-time task-oriented parsing. 

\paragraph{Evaluation scale.} To investigate our methods to a larger scale with more variances, we randomly choose another five predefined slot categories and collecting five-to-ten new slot values per category. To evaluate on new slot values, we build a set of modified TOP tests by randomly replacing the original values. We gradually increase this randomness to involve a larger percent of modifications on the original TOP test set (from 19.6\% to 37.3\%).
Figure \ref{fig-large-scale} shows that our lexicon-injected parsers are still able to achieve stable performances to tackle unseen slot values from varying modified test sets. Oppositely, the performances of span-based or generative parsers without lexicon-injected gradually decrease with more modifications involved.

\paragraph{Ablation study.} We further evaluate our models when new slot values are removed from the lexicon, as well as the effect of generalized representation on lexicon-injected parsers. As Appendix \ref{appendix_ablation_study} indicates, even without updating the lexicon, our lexicon-injected parsers still achieve comparable performance to state-of-the-art alternatives.

\subsection{Slot Disambiguation}
\label{slot_disambiguation}
\begin{table} \normalsize
\centering
\begin{tabular}{lr}
\hline \textbf{TOP Dataset} & \textbf{Accuracy} \\ \hline
Dev-matched set\dag & 98.42 \\
Test-matched set\dag & 98.26 \\
Modified test-matched set\dag & 98.24 \\
Modified test-matched set\S & 98.22 \\
\hline
\end{tabular}
\caption{\label{result-table-4} Accuracy of binary classification in slot disambiguation. Each set includes all matched span entries of comparing utterances to a lexicon. \dag\ and \S\ denote using the initial lexicon and the updated lexicon with new slot entries respectively.}
\end{table}

Table~\ref{result-table-4} describes the accuracy of our slot disambiguation technique. Overall, our slot disambiguation performs fairly well on different test sets. It demonstrates a promising ability to remove inappropriate slot match entries in the given context of utterances, thus reducing noises in subsequent parsing models. Logically, even if the disambiguation model fails to remove negative match entries, the later parser still holds the possibility to save it and produces a correct parsing result.

Once we incorporate new slot values into a modified test set, this binary classifier still achieves comparable performances. 
Results demonstrate that our slot disambiguation model is able to deal with these newly introduced span mismatches in given contexts, which may also be promising for a wide use in other NLP tasks such as named-entity recognition \citep{lample-ner-neural}. 

\section{Conclusion}
Overall, our novel lexicon-injected method to semantic parsing takes one more step towards real-time task-oriented parsing. Our best parser sets a new state of the art on the TOP dataset, with the adaptability of quickly handling unseen slot values. We provide a new solution for dialog systems (e.g., phone assistants), where a couple of dynamic slot lexicon tables are used to guide on identifying unseen slot entries. 

\section{Limitation} 
One limitation of our lexicon-injected model is that re-training is still needed if a novel slot category is introduced. It's indeed a difficult problem for the majority of models, given the fact that the last classifier layer is not even aware of the novel categories. However, it is not as frequent as introducing new slot values to a predefined slot category in real dialog systems. Besides, our parser has a CKY-style bottom-up decoder using dynamic programming. It can achieve globally optimal results in decoding and allow parent span to fetch bottom-up information from child span. However, it has a slightly slower performance O($N^3$) for an utterance of length N. 

In addition, we didn't use BART \citep{lewis2019bart} as a contextualized encoder in our proposed parsers given the computational expenses. In fact, our lexicon-injected parsers with RoBERTa have already achieved a better performance than decoupled BART \citep{sota}.

\section{Future work}
We believe such a lexicon-injected method can also be adopted in generative parsing models \citep{sota}, while we leave this validation in future work. We aim to speed up the inference by adapting and validating top-down greedy decoders. With the use of an additional lexicon, we believe our lexicon-injected models require less training data than common span-based parsers, while still achieving comparable performances. We will explore the few-shot semantic parsing \citep{chen2020low} using our lexicon-based methods. We would like evaluate on how the size of lexicon catalog impacts the performances of our models. There are additional datasets such as SNIPS \cite{DBLP:journals/corr/abs-1805-10190} and ATIS \cite{price1990evaluation} we would like to evaluate on in future work. However, the TOP dataset contains a deeper tree structure with nested intents and slots, which shows the value of our methods in an even more tricky setting.

Our proposed span-based parser and slot disambiguation model are not necessarily coupled. The two components can be retrained, compressed or accelerated separately, according to the real need. We believe such lexicon-injected method can also be adopted in generative parsing models \citep{sota}, while we leave this validation in future work.



\clearpage
\bibliographystyle{acl_natbib}
\bibliography{anthology}

\begin{thebibliography}{31}
\expandafter\ifx\csname natexlab\endcsname\relax\def\natexlab#1{#1}\fi

\bibitem[{Aghajanyan et~al.(2020)Aghajanyan, Maillard, Shrivastava, Diedrick,
  Haeger, Li, Mehdad, Stoyanov, Kumar, Lewis, and Gupta}]{sota}
Armen Aghajanyan, Jean Maillard, Akshat Shrivastava, Keith Diedrick, Michael
  Haeger, Haoran Li, Yashar Mehdad, Veselin Stoyanov, Anuj Kumar, Mike Lewis,
  and Sonal Gupta. 2020.
\newblock Conversational semantic parsing.
\newblock \emph{In Empirical Methods in Natural Language Processing (EMNLP)}.

\bibitem[{Berant et~al.(2013)Berant, Chou, Frostig, and
  Liang}]{berant2013semantic}
Jonathan Berant, Andrew Chou, Roy Frostig, and Percy Liang. 2013.
\newblock Semantic parsing on freebase from question-answer pairs.
\newblock In \emph{In Empirical Methods in Natural Language Processing
  (EMNLP)}.

\bibitem[{Black et~al.(1991)Black, Abney, Flickinger, Gdaniec, Grishman,
  Harrison, Hindle, Ingria, Jelinek, Klavans et~al.}]{black1991procedure}
Ezra Black, Steven Abney, Dan Flickinger, Claudia Gdaniec, Ralph Grishman, Phil
  Harrison, Donald Hindle, Robert Ingria, Frederick Jelinek, Judith~L Klavans,
  et~al. 1991.
\newblock A procedure for quantitatively comparing the syntactic coverage of
  english grammars.
\newblock In \emph{Speech and Natural Language: Proceedings of a Workshop Held
  at Pacific Grove, California, February 19-22, 1991}.

\bibitem[{Chen et~al.(2020)Chen, Ghoshal, Mehdad, Zettlemoyer, and
  Gupta}]{chen2020low}
Xilun Chen, Asish Ghoshal, Yashar Mehdad, Luke Zettlemoyer, and Sonal Gupta.
  2020.
\newblock Low-resource domain adaptation for compositional task-oriented
  semantic parsing.
\newblock \emph{arXiv preprint arXiv:2010.03546}.

\bibitem[{Coucke et~al.(2018)Coucke, Saade, Ball, Bluche, Caulier, Leroy,
  Doumouro, Gisselbrecht, Caltagirone, Lavril, Primet, and
  Dureau}]{DBLP:journals/corr/abs-1805-10190}
Alice Coucke, Alaa Saade, Adrien Ball, Th{\'{e}}odore Bluche, Alexandre
  Caulier, David Leroy, Cl{\'{e}}ment Doumouro, Thibault Gisselbrecht,
  Francesco Caltagirone, Thibaut Lavril, Ma{\"{e}}l Primet, and Joseph Dureau.
  2018.
\newblock \href {http://arxiv.org/abs/1805.10190} {Snips voice platform: an
  embedded spoken language understanding system for private-by-design voice
  interfaces}.
\newblock \emph{CoRR}, abs/1805.10190.

\bibitem[{Devlin et~al.(2019)Devlin, Chang, Lee, and Toutanova}]{bert}
Jacob Devlin, Ming-Wei Chang, Kenton Lee, and Kristina Toutanova. 2019.
\newblock {BERT}: Pre-training of deep bidirectional transformers for language
  understanding.
\newblock \emph{In North American Association for Computational Linguistics:
  Human Language Technologies (NAACL-HLT).}

\bibitem[{Einolghozati et~al.(2019)Einolghozati, Pasupat, Gupta, Shah, Mohit,
  Lewis, and Zettlemoyer}]{einolghozati2019improving}
Arash Einolghozati, Panupong Pasupat, Sonal Gupta, Rushin Shah, Mrinal Mohit,
  Mike Lewis, and Luke Zettlemoyer. 2019.
\newblock Improving semantic parsing for task oriented dialog.
\newblock \emph{In Conversational AI Workshop at NeurIPS.}

\bibitem[{Gaddy et~al.(2018)Gaddy, Stern, and Klein}]{gaddy2018s}
David Gaddy, Mitchell Stern, and Dan Klein. 2018.
\newblock What's going on in neural constituency parsers? an analysis.
\newblock \emph{In North American Association for Computational Linguistics:
  Human Language Technologies (NAACL-HLT)}.

\bibitem[{Gupta et~al.(2018)Gupta, Shah, Mohit, Kumar, and
  Lewis}]{gupta2018semantic}
Sonal Gupta, Rushin Shah, Mrinal Mohit, Anuj Kumar, and Mike Lewis. 2018.
\newblock Semantic parsing for task oriented dialog using hierarchical
  representations.
\newblock \emph{In Empirical Methods in Natural Language Processing (EMNLP)}.

\bibitem[{Herzig and Berant(2020)}]{herzig2020span}
Jonathan Herzig and Jonathan Berant. 2020.
\newblock Span-based semantic parsing for compositional generalization.
\newblock \emph{arXiv preprint arXiv:2009.06040}.

\bibitem[{Jia and Liang(2016)}]{jia-liang-2016}
Robin Jia and Percy Liang. 2016.
\newblock Data recombination for neural semantic parsing.
\newblock In \emph{In Association for Computational Linguistics (ACL)}.

\bibitem[{Kipper et~al.(2000)Kipper, Dang, Palmer et~al.}]{kipper2000class}
Karin Kipper, Hoa~Trang Dang, Martha Palmer, et~al. 2000.
\newblock Class-based construction of a verb lexicon.
\newblock \emph{AAAI/IAAI}.

\bibitem[{Kitaev and Klein(2018)}]{kitaev2018constituency}
Nikita Kitaev and Dan Klein. 2018.
\newblock Constituency parsing with a self-attentive encoder.
\newblock \emph{In Association for Computational Linguistics (ACL)}.

\bibitem[{Lample et~al.(2016)Lample, Ballesteros, Subramanian, Kawakami, and
  Dyer}]{lample-ner-neural}
Guillaume Lample, Miguel Ballesteros, Sandeep Subramanian, Kazuya Kawakami, and
  Chris Dyer. 2016.
\newblock Neural architectures for named entity recognition.
\newblock In \emph{Proceedings of the 2016 Conference of the North {A}merican
  Chapter of the Association for Computational Linguistics: Human Language
  Technologies}.

\bibitem[{Lewis et~al.(2019)Lewis, Liu, Goyal, Ghazvininejad, Mohamed, Levy,
  Stoyanov, and Zettlemoyer}]{lewis2019bart}
Mike Lewis, Yinhan Liu, Naman Goyal, Marjan Ghazvininejad, Abdelrahman Mohamed,
  Omer Levy, Ves Stoyanov, and Luke Zettlemoyer. 2019.
\newblock Bart: Denoising sequence-to-sequence pre-training for natural
  language generation, translation, and comprehension.
\newblock \emph{arXiv preprint arXiv:1910.13461}.

\bibitem[{Li et~al.(2020)Li, Yan, Qiu, and Huang}]{li2020flat}
Xiaonan Li, Hang Yan, Xipeng Qiu, and Xuanjing Huang. 2020.
\newblock Flat: Chinese ner using flat-lattice transformer.
\newblock \emph{In Association for Computational Linguistics (ACL)}.

\bibitem[{Liu et~al.(2019)Liu, Ott, Goyal, Du, Joshi, Chen, Levy, Lewis,
  Zettlemoyer, and Stoyanov}]{liu2019roberta}
Yinhan Liu, Myle Ott, Naman Goyal, Jingfei Du, Mandar Joshi, Danqi Chen, Omer
  Levy, Mike Lewis, Luke Zettlemoyer, and Veselin Stoyanov. 2019.
\newblock Roberta: A robustly optimized bert pretraining approach.
\newblock \emph{arXiv preprint arXiv:1907.11692}.

\bibitem[{Louvan and Magnini(2020)}]{louvan-magnini-2020}
Samuel Louvan and Bernardo Magnini. 2020.
\newblock Recent neural methods on slot filling and intent classification for
  task-oriented dialogue systems: A survey.
\newblock \emph{In Proceedings of the 28th International Conference on
  Computational Linguistics}.

\bibitem[{Miller(1995)}]{miller1995wordnet}
George~A Miller. 1995.
\newblock Wordnet: a lexical database for english.
\newblock \emph{Communications of the ACM}.

\bibitem[{Pasupat et~al.(2019)Pasupat, Gupta, Mandyam, Shah, Lewis, and
  Zettlemoyer}]{pasupat2019span}
Panupong Pasupat, Sonal Gupta, Karishma Mandyam, Rushin Shah, Mike Lewis, and
  Luke Zettlemoyer. 2019.
\newblock Span-based hierarchical semantic parsing for task-oriented dialog.
\newblock \emph{In Empirical Methods in Natural Language Processing (EMNLP)}.

\bibitem[{Price(1990)}]{price1990evaluation}
Patti Price. 1990.
\newblock Evaluation of spoken language systems: The atis domain.
\newblock In \emph{Speech and Natural Language: Proceedings of a Workshop Held
  at Hidden Valley, Pennsylvania}.

\bibitem[{Rongali et~al.(2020)Rongali, Soldaini, Monti, and
  Hamza}]{rongali2020don}
Subendhu Rongali, Luca Soldaini, Emilio Monti, and Wael Hamza. 2020.
\newblock Don’t parse, generate! a sequence to sequence architecture for
  task-oriented semantic parsing.
\newblock In \emph{Proceedings of The Web Conference 2020}, pages 2962--2968.

\bibitem[{Shi and Mihalcea(2005)}]{shi2005putting}
Lei Shi and Rada Mihalcea. 2005.
\newblock Putting pieces together: Combining framenet, verbnet and wordnet for
  robust semantic parsing.
\newblock In \emph{International conference on intelligent text processing and
  computational linguistics}. Springer.

\bibitem[{Stern et~al.(2017)Stern, Andreas, and Klein}]{stern2017minimal}
Mitchell Stern, Jacob Andreas, and Dan Klein. 2017.
\newblock A minimal span-based neural constituency parser.
\newblock \emph{In Association for Computational Linguistics (ACL)}.

\bibitem[{Taboada et~al.(2011)Taboada, Brooke, Tofiloski, Voll, and
  Stede}]{taboada2011lexicon}
Maite Taboada, Julian Brooke, Milan Tofiloski, Kimberly Voll, and Manfred
  Stede. 2011.
\newblock Lexicon-based methods for sentiment analysis.
\newblock \emph{Computational linguistics}.

\bibitem[{Tur and De~Mori(2011)}]{tur2011spoken}
Gokhan Tur and Renato De~Mori. 2011.
\newblock \emph{Spoken language understanding: Systems for extracting semantic
  information from speech}.
\newblock John Wiley \& Sons.

\bibitem[{Van~Noord et~al.(2018)Van~Noord, Abzianidze, Toral, and
  Bos}]{van2018exploring}
Rik Van~Noord, Lasha Abzianidze, Antonio Toral, and Johan Bos. 2018.
\newblock Exploring neural methods for parsing discourse representation
  structures.
\newblock \emph{Transactions of the Association for Computational Linguistics}.

\bibitem[{Vaswani et~al.(2017)Vaswani, Shazeer, Parmar, Uszkoreit, Jones,
  Gomez, Kaiser, and Polosukhin}]{vaswani2017attention}
Ashish Vaswani, Noam Shazeer, Niki Parmar, Jakob Uszkoreit, Llion Jones,
  Aidan~N Gomez, {\L}ukasz Kaiser, and Illia Polosukhin. 2017.
\newblock Attention is all you need.
\newblock \emph{Advances in neural information processing systems}, pages
  5998--6008.

\bibitem[{Williams et~al.(2016)Williams, Raux, and
  Henderson}]{williams2016dialog}
Jason~D Williams, Antoine Raux, and Matthew Henderson. 2016.
\newblock The dialog state tracking challenge series: A review.
\newblock \emph{Dialogue \& Discourse}.

\bibitem[{Xu and Hu(2018)}]{xu-hu-2018}
Puyang Xu and Qi~Hu. 2018.
\newblock An end-to-end approach for handling unknown slot values in dialogue
  state tracking.
\newblock In \emph{In Association for Computational Linguistics (ACL)}.

\bibitem[{Zhu and Yu(2017)}]{zhu2017encoder}
Su~Zhu and Kai Yu. 2017.
\newblock Encoder-decoder with focus-mechanism for sequence labelling based
  spoken language understanding.
\newblock In \emph{2017 IEEE International Conference on Acoustics, Speech and
  Signal Processing (ICASSP)}. IEEE.

\end{thebibliography}
\appendix
\section{Training parameters}
\label{appendix_A}
The encoder is a 2-layer transformer with hidden size 512 and 8 attentions heads as in \citet{bert}. In the base model, We use 512-dimensional embeddings for $w_{i}$, $p_{i}$, and thus the dimension of $x_{i}$ is 1024. In the lexicon-injected parser, we concatenate another $q_{i}$ with the 128-dimensional embeddings as descirbed in section \ref{lexion_model}. We experimented with variants of input embeddings and were unaware of any significant difference in parsing performances. Overall, we performed five trials on each task with with multiple settings of random seeds, and averaged resulting performances.

In addition, we randomly split the output $h_i$ from the word-based encoder described in section \ref{encoder} to form boundary representation, because our preliminary study found the way of splitting makes no difference. It shows that the encoder-decoder architecture is able to adapt to a particular way of splitting $h_i$ to form boundary representation.



\begin{table}[t!]
\centering
\begin{tabular}{lrl}
\hline \textbf{Slot Category} & \textbf{Value} \\ \hline
\emph{SL:TYPE\_RELATION} & brother in law \\
 & roommate \\
 & homie \\
 & stepfather \\
 & boy friend \\
 & \\
 \emph{SL:LOCATION} & beach park \\
 & school building \\
 & street - 25 \\
 & bridge \\
 & rocket company \\
 & \\
 \emph{SL:POINT\_ON\_MAP} & Silicon Valley \\
 & Red Rock Canyon \\
 & White House \\
 & Ahaggar National Park \\
 & Singapore \\
\hline
\end{tabular}
\caption{\label{slot_example} Examples of new slot values designed for modifying TOP test set.}
\end{table}

\section{Modified TOP test set}
\label{appendix_test_set}
Table \ref{slot_example} provides some examples of our randomly collected new slot values for updating the lexicon table. We aim to collect unseen values that are meaningfully consistent with the corresponding predefined slot category. These new slot values were originally missing in the train set, but are very likely to appear in real user utterances. Notice that we are unaware of whether a newly collected slot value is used to build modified test sets.

Updating the lexicon table is a common use case in real task-oriented dialog: the lexicon table serves like an external knowledge resource in the given task domain. General or domain slot resources in this task are keeping updating the lexicon via human effort or web crawler tools. For instance, the phone contact lexicon keeps updated and is used to guide personal virtual assistants. The recent popular songs are crawled from websites and periodically added to the music name lexicon for potential use by phone assistants. Notice that we use case-insensitive comparison to match utterances to the lexicon.

\section{Supplementary Experiments}
\subsection{Oracle parser}
\label{appendix_oracle_parser}
The oracle parser serves as to determine the upper bound (\textbf{95.26\%}) for the lexicon-injected gains, as shown in Table \ref{oracle-parser}. This parser always picks the correct slot-category for each span via the ground-truth dev trees. This result suggests there exists considerable room for improving the lexicon-injected parser, which motivates us to come up with a slot disambiguation technique. In addition, it shows that a better slot disambiguation technique is able to further improve the paring performance.

\begin{table}[t!]
\centering
\begin{tabular}{lrl}
\hline \textbf{Method} & \textbf{Acc} & \textbf{F1} \\ \hline
\textbf{Our lexicon-injected parser:} &  &  \\
w/ Oracle (dev)\dag & 94.52 & 98.52 \\
w/ Oracle + GR (dev)\dag & 95.26 & 98.86 \\
\hline
\end{tabular}
\caption{\label{oracle-parser} Performances of our oracle parsers. GR and \dag\ denote generalized representation and use BERT-base model, respectively.}
\end{table}

\begin{table}
\centering
\begin{tabular}{lrl}
\hline \textbf{Method} & \textbf{Acc} & \textbf{F1} \\ \hline
\textbf{Using the up-to-date lexicon:} &  &  \\
w/ Slot Disambiguation\dag & 84.80 & 95.92 \\
w/ SD + GR\dag & 86.30 & 96.25 \\
w/ SD + GR + RoBERTa & \textbf{87.27} & \textbf{96.54} \\
\textbf{Using the original lexicon:} &  &  \\
w/ Slot Disambiguation\dag & 71.75 & 93.23 \\
w/ SD + GR\dag & 70.95 & 93.19 \\
w/ SD + GR + RoBERTa & 71.87 & 93.49 \\
w/ SD + RoBERTa & 72.78 & 93.73 \\
\hline
\end{tabular}
\caption{\label{ablation_study} Ablation study - comparison of lexicon-injected parsers when using the up-to-date lexicon or the original lexicon to parse the modified TOP test set. SD, GR and \dag\ denote slot disambiguation, generalized representation and using BERT-base respectively.}
\end{table}

\subsection{Ablation Study}
\label{appendix_ablation_study}
We believe the updated lexicon, like a knowledge base with new knowledge updated, serves as a main factor to help on tackling with unseen slot values. Therefore, we evaluate the model performance when these new values are removed from the lexicon, \textit{i.e.,} using the original lexicon to guide our models on paring the same modified test set. 

As shown in Table~\ref{ablation_study}, the performance decrease in accuracy also applies to our lexicon-injected parsers, since newly added slot values are not found in the original lexicon. 

Results show that even without human intervention on updating the lexicon, lexicon-injected parsers still achieve comparable performance to state-of-the-art non-lexicon-injected parsers in Table~\ref{modified-test-result}.
It shows the promises of our model in real dialog systems, with its adaptability to dynamic lexicons and embracing unknowns in real-time.

\paragraph{Generalized Representation.}
In Table~\ref{ablation_study}, our lexicon-injected model (SD + GR) is not superior to the model with SD only when we use the original lexicon. The decrease in match accuracy is \textbf{0.8\%} in BERT-based and \textbf{0.91\%} in RoBERTa-based models. However, it leads to an improvement (\textbf{+1.5\%}) after we use the updated lexicon as in Table~\ref{modified-test-result}.

This result is beyond doubt because GR abandons external word embedding and lies on slot-category embedding for slot entries in the lexicon. This method enforces slot-category to weight in the same way as word embedding, meanwhile, to represent a group of slot values within the same category. It empowers the weight of lexical information on parsing decisions. It thus brings a noticeable improvement when we have adequate lexical information. However, it's a trade-off in use that GR lightly suffers from the case when lexical information is not imported yet.

\end{document}